\documentclass[10pt,twocolumn,apaper]{article}
\usepackage[pagenumbers]{wacv} % To force page numbers, e.g. for an arXiv version

\usepackage{graphicx}
\usepackage{amsmath}
\usepackage{amssymb}
\usepackage{booktabs}

\usepackage[normalem]{ulem}
\useunder{\uline}{\ul}{}      % 230119

\usepackage{bbding}
\usepackage{pifont}
\usepackage{wasysym}
\usepackage{amssymb}

\usepackage{multirow}

\usepackage[misc]{ifsym}   % 220712, corresponding 

\usepackage[accsupp]{axessibility} % 231026, Improves PDF readability for those with disabilities.

\usepackage[pagebackref,breaklinks,colorlinks]{hyperref}

\usepackage[capitalize]{cleveref}
\crefname{section}{Sec.}{Secs.}
\Crefname{section}{Section}{Sections}
\Crefname{table}{Table}{Tables}
\crefname{table}{Tab.}{Tabs.}

\def\wacvPaperID{1090} % *** Enter the WACV Paper ID here
\def\confName{WACV}
\def\confYear{2024}

\begin{document}

\title{Understanding Dark Scenes by Contrasting Multi-Modal Observations}

% \author{Xiaoyu Dong\\
% The University of Tokyo, RIKEN AIP\\
% Tokyo, Japan\\  % 占空间
% {\tt\small dong@ms.k.u-tokyo.ac.jp}
% \and
% Naoto Yokoya\\  %$^{(\textrm{\Letter})}$   % 通讯作者
% The University of Tokyo, RIKEN AIP\\
% Tokyo, Japan\\  % 占空间
% {\tt\small yokoya@k.u-tokyo.ac.jp}
% }
\author{Xiaoyu Dong$^{1,2}$ and Naoto Yokoya$^{1,2,\textrm{\Letter}}$\\  % (\textrm{\Letter})
$^{1}$The University of Tokyo, Japan\\
$^{2}$RIKEN AIP, Japan\\  %
{\tt\small dong@ms.k.u-tokyo.ac.jp, yokoya@k.u-tokyo.ac.jp}
}
\maketitle

%%%%%%%%% ABSTRACT
\begin{abstract}
Understanding dark scenes based on multi-modal image data is challenging, as both the visible and auxiliary modalities provide limited semantic information for the task. Previous methods focus on fusing the two modalities but neglect the correlations among semantic classes when minimizing losses to align pixels with labels, resulting in inaccurate class predictions. To address these issues, we introduce a supervised multi-modal contrastive learning approach to increase the semantic discriminability of the learned multi-modal feature spaces by jointly performing cross-modal and intra-modal contrast under the supervision of the class correlations. The cross-modal contrast encourages same-class embeddings from across the two modalities to be closer and pushes different-class ones apart. The intra-modal contrast forces same-class or different-class embeddings within each modality to be together or apart. We validate our approach on a variety of tasks that cover diverse light conditions and image modalities. Experiments show that our approach can effectively enhance dark scene understanding based on multi-modal images with limited semantics by shaping semantic-discriminative feature spaces. Comparisons with previous methods demonstrate our state-of-the-art performance. 
Code and pretrained models are available at~\url{https://github.com/palmdong/SMMCL}. % put plain text in cmt

\end{abstract}

%%%%%%%%% BODY TEXT
\section{Introduction}
\label{sec:intro}

\begin{figure}
  \centering
  \begin{subfigure}{0.49\linewidth}
    \includegraphics[width=1\linewidth]{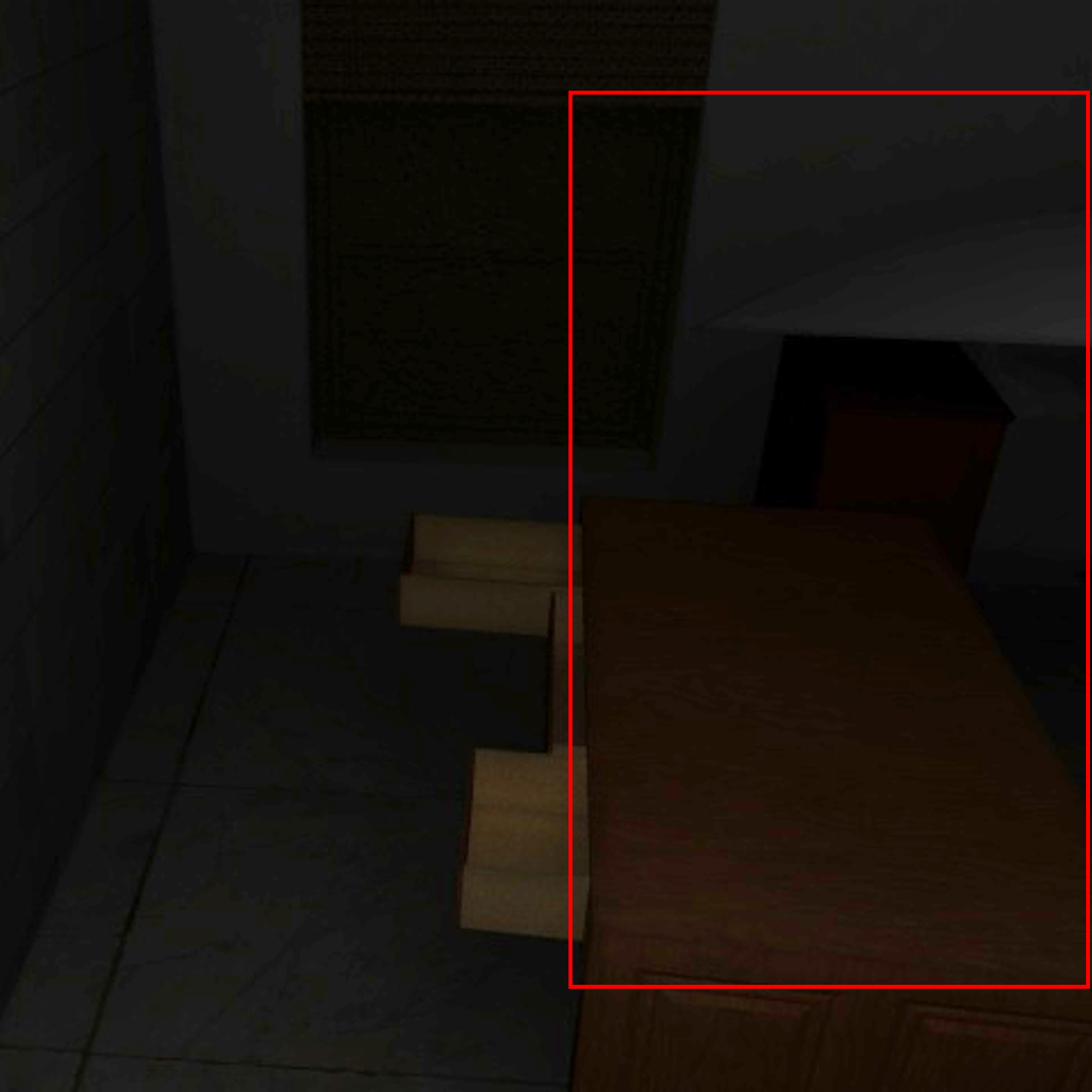}
    \caption{RGB}
    % \label{fig1_a}
  \end{subfigure}
  \begin{subfigure}{0.49\linewidth}
    \includegraphics[width=1\linewidth]{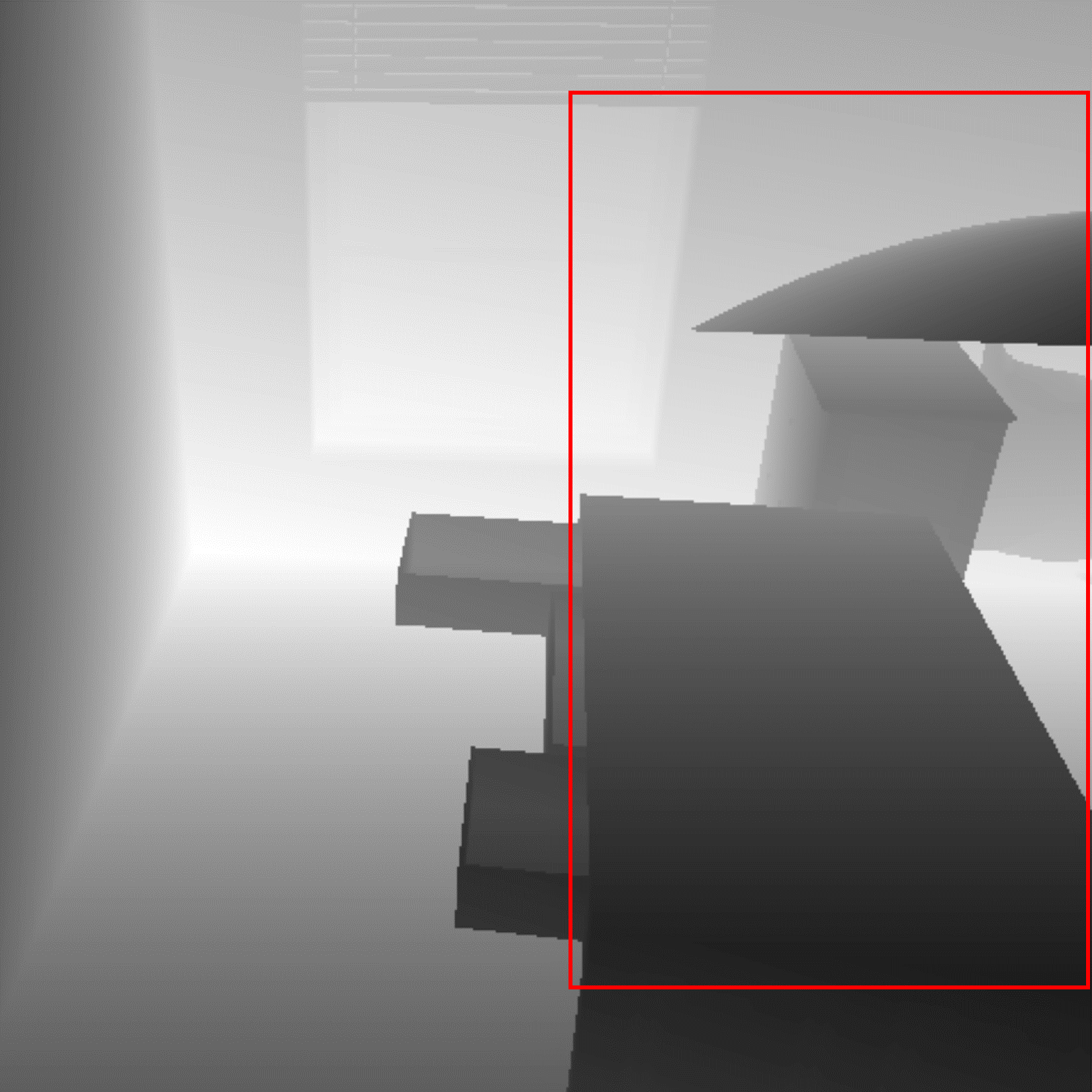}
    \caption{Depth}
  \end{subfigure}
  % \hfill  % 填充空白
  \begin{subfigure}{0.24\linewidth}
    \includegraphics[width=1\linewidth]{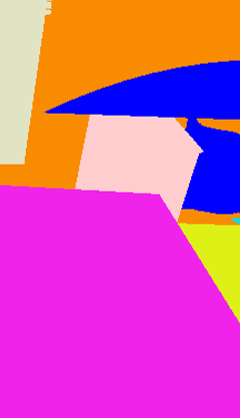}
    \caption{Ground Truth}
  \end{subfigure}
  \begin{subfigure}{0.24\linewidth}
    \includegraphics[width=1\linewidth]{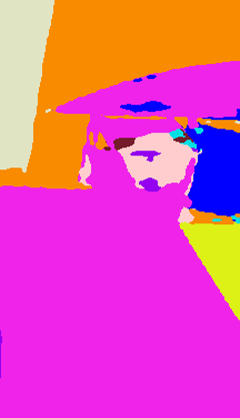}
    \caption{CEN~\cite{pami22_cen}}
  \end{subfigure}
  \begin{subfigure}{0.24\linewidth}
    \includegraphics[width=1\linewidth]{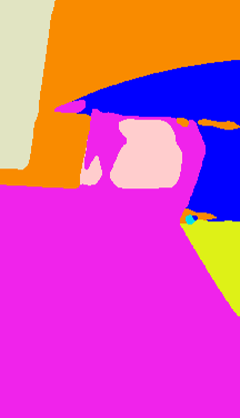}
    \caption{ToFusion~\cite{cvpr22_token}}  
  \end{subfigure}
  \begin{subfigure}{0.24\linewidth}
    \includegraphics[width=1\linewidth]{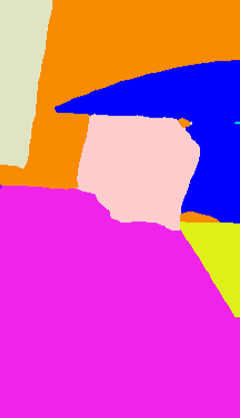}
    \caption{Ours}
  \end{subfigure}
  \caption{Low-light indoor scene segmentation from RGB-depth data. 
   Compared to state-of-the-art methods, our model 
   with supervised multi-modal contrastive learning achieves higher accuracy.
   }
  \label{fig1}
\end{figure}

% dark scene semantic segmentation
A robust scene understanding capability in dark environments, including low-light indoor and nighttime outdoor environments, is important to automated work systems such as indoor robots and automotive vehicles~\cite{cui_iccv21_detection,cvpr22_shift}. 
However, semantic segmentation on dark scenes, especially based on observation images from visible RGB modality, is not trivial due to the poor visibility of spatial content in images caused by adverse light conditions~\cite{cui_2023_nerf,cvpr22_nightlab}. 
% light conditions, dark environments 

% multi-modal image semantic segmentation
Combining images from multiple modalities that can provide complementary spatial information for a scene, often visible RGB modality and an auxiliary depth or thermal modality, has been proved beneficial to semantic segmentation tasks~\cite{eccv12_nyu,iros17_mfnet}. 
And numerous multi-modal image semantic segmentation methods have been developed~\cite{cvpr22_token,pami22_cen,cvpr21_abmdernet,iccv21_shape,eccv20_sagate,aaai22_egfnet,iros20_heat}.

% 数据问题
However, in the task of dark scene segmentation, the visible and auxiliary modalities 
both provide limited semantic information. 
To be specific: 
The visible modality reflects contextual semantic cues in RGB color space, but available cues are usually limited due to its dark nature~\cite{cvpr22_nightlab}. 
The auxiliary modality is robust to adverse lights and can provide rich geometry cues for dark environments, but is lacking in contextual semantics~\cite{iccv21_shape,eccv20_sagate}. 
These cause low discrimination between different semantic classes, as shown in~\cref{fig1}. % 0308, 类别间的discrimination/hanzhehu p7, 特征空间的semantic disciminability 
% 方法问题
Previous multi-modal image segmentation methods~\cite{2022_cmx,cvpr22_token,pami22_cen,cvpr21_abmdernet,iccv21_shape,eccv20_sagate} focus on developing fusion techniques to combine the two modalities, 
then minimizing cross-entropy losses to align pixels with corresponding labels without considering the correlations (similarities and differences) among semantic classes. 
As a result, they tend to predict inaccurate class information for objects in darkness~(\cref{fig1}). 
Overall, multi-modal dark scene understanding remains an open problem.

\begin{figure*}[t]
  \centering
  \includegraphics[width=0.86\linewidth]{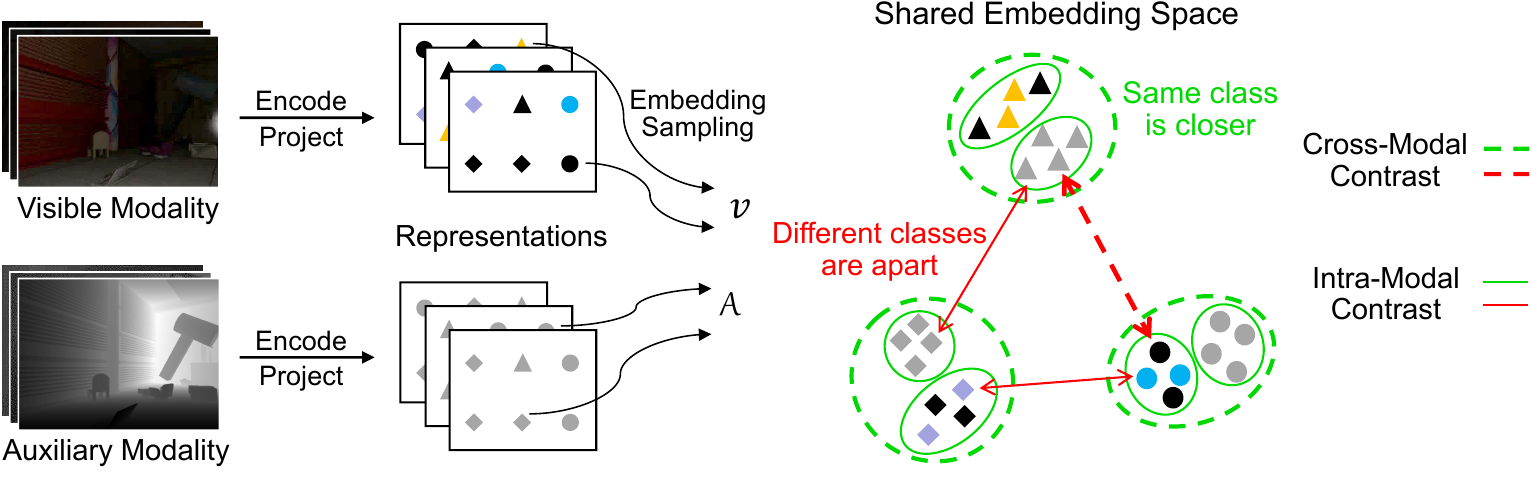} 
  \caption{
  An illustration of our supervised multi-modal contrastive learning approach. 
  During training, embeddings from the visible and auxiliary modalities are cast to a shared space, where cross-modal and intra-modal contrast are jointly performed under the supervision of the class correlations.  
  Same shape means the embeddings are from the same semantic class and are positive to each other.
  Colored, black, and grey mean the embeddings carry semantic cues, are not observable, and lack contextual semantics, respectively.
  } 
  \label{fig2_mmcl}
\end{figure*}

% 描述和图匹配吗?
% cl操作在embedding space, 但shape的是encoder学的feature space
In this paper, we address the issues by increasing the semantic discriminability of the learned feature spaces via contrastive learning.
Specifically, we introduce a supervised multi-modal contrastive learning approach~(\cref{fig2_mmcl}) to boost the learning on the visible and auxiliary modalities and encourage them to be semantic-discriminative, by jointly performing cross-modal and intra-modal contrast under the supervision of the class correlations. 
The cross-modal contrast encourages same-class embeddings from across the two modalities to be closer and simultaneously pushes different-class ones apart.  % 再突出同类不同类?
Within each modality, the intra-modal contrast pulls together embeddings from the same class and forces apart those from different classes. 
By regularizing the embeddings with considering the class similarities and differences, the encoder feature spaces learned from the two modalities can show higher semantic discriminability. % 上面讲了correlations指similarities和differences
With the adoption of our approach, our segmentation model achieves much higher accuracy when understanding dark scenes based on multi-modal images with limited semantics~(\cref{fig1}).

\textbf{Contributions:}
(1)~We tackle dark scene understanding from a new perspective by contrasting multi-modal images with limited semantics.
(2)~We introduce the first supervised multi-modal contrastive learning approach for image segmentation,
and show it can effectively enhance dark scene understanding by shaping semantic-discriminative feature spaces. 
(3)~We validate our approach on low-light, nighttime, and normal-light conditions, 
indoor and outdoor scenes, 
and RGB, depth, and thermal modalities,
demonstrating its effectiveness, generalizability, and applicability. 
(4)~We compare our model and approach with state-of-the-art methods on different tasks, showing our superiority quantitatively and qualitatively.

%-------------------------------------------------------------------------
%-------------------------------------------------------------------------

\section{Related Work}  

\subsection{Semantic Segmentation}  
Semantic segmentation is the task of understanding scenes by assigning each pixel in an image to a specific class. 
Since FCN~\cite{cvpr15_fcn} was proposed, numerous CNN-based semantic segmentation methods have been developed. 
Representative work includes 
the DeepLab series~\cite{pami17_deeplab,arxiv17_deeplab,eccv18_deeplab}, multi-scale networks~\cite{cvpr17_pspnet,cvpr19_hrnet,pami20_hrnet}, % nightlab 
boundary or context-aware networks~\cite{cvpr16_boundary,iccv19_boundary,cvpr20_boundary,cvpr18_context,cvpr19_context,cvpr20_context}, and attention-based networks~\cite{eccv18_att,cvpr19_att,cvpr20_att,pami20_att}. % segformer
Most recently, Vision Transformers~\cite{iclr23_vitadapter,cvpr22_mpvit,nips22_segnext,nips_21segformer,nips21_hrformer,iccv21_segmentor} have shown great potential and outperformed CNN-based methods. 
However, these advances are made for normal-light scenarios. 
In practical applications, 
there is a need for a robust scene understanding capability in dark environments. 

%------------------------------------------------------------------------

\subsection{Dark Scene Semantic Segmentation}  
Existing dark scene semantic segmentation methods are mainly developed based on visible RGB data, and can be divided into unsupervised domain adaptation methods and supervised methods. % 主要是单模态, visible RGB. mainly因为rgbt系列 xmuda等
Unsupervised domain adaptation methods~\cite{cvpr23_uda,cvpr22_distillation,cvpr21_onestage,pami20_guided,iccv19_guided,itsc18_dai} tackle unlabeled dark scenes by transferring knowledge from labeled normal-light scenes that share similar spatial content. % 0306, objects
The problems with such methods are that they require paired dark-normal training data, which is hard to collect in practice, and their unsupervised working piepline causes limited performance~\cite{cvpr22_nightlab,pami22_xmuda}.
Supervised methods~\cite{cvpr22_nightlab,arxiv22_nightfilters,arxiv22_nightfrequency,tip21_nightCity,2021_llrgbd,2022_llrgbd} learn the task from labeled dark scene data directly, and so avoid the need for additional normal-light data. 
However, they still show unsatisfactory performance on regions of poor visibility because reliable contextual cues in the visible modality is limited~\cite{cvpr22_nightlab,2022_cmx}.
Therefore, recent methods~\cite{pami22_xmuda,2022_cmx,cvpr21_abmdernet,iccv21_sparsedense} combine auxiliary modalities that can provide robust geometry cues for even dark environments.

%------------------------------------------------------------------------

\subsection{Multi-Modal Image Semantic Segmentation} 
Multi-modal image data, \textit{e.g.}, RGB-depth and RGB-thermal, has been proven beneficial to semantic segmentation due to the capability of providing complementary spatial information for scenes.
Numerous methods with advanced fusion techniques, such as token fusion~\cite{cvpr22_token}, channel exchanging~\cite{pami22_cen,nips20_cen}, feature interaction modules~\cite{cvpr21_abmdernet,2022_cmx,eccv20_sagate,aaai22_egfnet,eccv22_uctnet}, and novel convolutions~\cite{iccv21_shape,eccv20_25conv,eccv18_depthconv,tip21_sconv}, have been developed and show promising performance, especially for normal-light scenes. 
On dark scenes, however, they still suffer inaccurate class predictions because:~(1)~The visible and auxiliary modalities both provide limited semantic information, which causes low discrimination between different classes. (2)~They neglect the correlations among classes when minimizing losses to align pixels with labels.
%
% 0228, 它们without considering the correlation among semantic classes 
To address the issues, we introduce a supervised multi-modal contrastive learning approach to increase the semantic discriminability of the learned feature spaces of the two modalities, by regularizing their embeddings under the supervision of the class correlations. 
We demonstrate that our approach enables a higher accuracy in understanding dark scenes and also generalizes well to normal-light scenes.

%------------------------------------------------------------------------

\subsection{Contrastive Learning} 

The idea of self-supervised contrastive learning~\cite{cvpr06_cl_yann,cvpr18_nonparametric,cvpr20_moco,arxiv20_simclr} is to pull an anchor closer to a positive sample in embedding space and further from many negative samples, without knowing their labels. 
Supervised contrastive learning~\cite{nips20_supervised} leverages label information to align embeddings and directly consider positive samples from the same class and negative classes from different classes.
The use of a supervised paradigm enables better generalization in general image classification and segmentation tasks~\cite{nips20_supervised,eccv22_mscs,iccv21_supervised_bank,iccv21_supervised_efficient,iccv2021_region,iccv21_supervised_cross}. % 这里只能引supervised

In the field of multi-modal learning, 
various self-supervised contrastive techniques~\cite{cvpr22_crosspoint,iccv21_crossclr,cvpr21_xmcgan,ral23_people,iros20_unirecognition,iccv21_tuple} have been presented.  
We introduce a supervised multi-modal learning approach to tackle dark scene understanding.
Unlike those self-supervised contrastive techniques~\cite{ral23_people,iros20_unirecognition,iccv21_tuple}, which need to generate positive and/or negative samples via complicated augmentation, our supervised paradigm effectively aligns multi-modal embeddings by leveraging available class labels. % align, 方法章说align as p n samples
This allows to directly and fully exploit the class correlations and the correspondence between cross-modal contextual and geometry cues. % 贯穿全文的是under the supervision of class correaltions, 另方法章说了use label to measure correlation
We demonstrate the effectiveness and superiority of our approach with comprehensive ablations and comparisons.

% 2-与其他跨模态cl不同：直接根据/利用class assignment对齐p n样本，regularize feature spaces（crosscl跨模态只考虑n） / 0222, 即supervised?
% 1-与其他有监督cl不同：多p多n，模态内且跨模态augment对比样本diversity，以让两种模态的cues…
% -或合一起分两条讲，3 experiments show..

%------------------------------------------------------------------------
%------------------------------------------------------------------------

\section{Method}

We first give an overview of our model, then detail our supervised multi-modal contrastive learning approach.

%------------------------------------------------------------------------

\begin{figure}[t]
  \centering
   \includegraphics[width=1\linewidth]{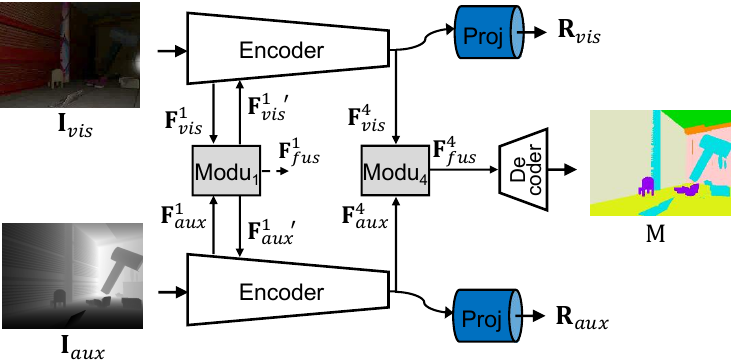}
   \caption{
   An illustration of our segmentation model.
   Final features from the encoders are mapped to representations by the projectors. 
   The representations are further utilized to generate embeddings in our supervised multi-modal contrastive learning approach.} 
   \label{fig3_model}
\end{figure}

% two encoders, feature processing modules, projectors, decoder, CE loss 
% describe "how they work" at a high level

\subsection{Model Overview} 
Our model is illustrated in~\cref{fig3_model}.  
Given a dark scene image $\textbf{I}_{vis}\in\mathbb{R}^{H\times W\times 3}$ in visible modality and its counterpart $\textbf{I}_{aux}\in\mathbb{R}^{H\times W}$ from an auxiliary modality, 
we use two encoders to encode them and extract multi-modal features 
$\textbf{F}_{vis}^m \in~\mathbb{R}^{h\times w\times c}$ and $\textbf{F}_{aux}^m \in~\mathbb{R}^{h\times w\times c}$, 
where $m$ = 1, 2, 3, 4 corresponds to the stage in the encoders. % $m$ = 1...4
Intermediate modules are developed to further process the features. % 'process' covers...

\begin{figure}[t]
  \centering
   \includegraphics[width=0.78\linewidth]{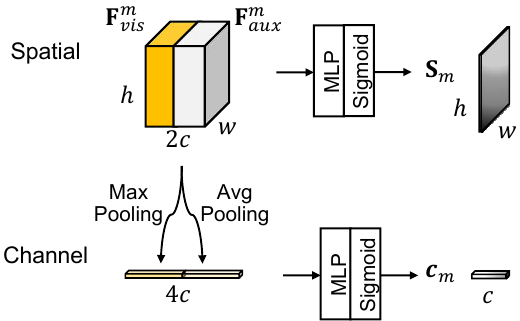}
   \caption{
   The spatial and channel coefficient learning in our intermediate modules.
   $\textbf{F}_{vis}^m$ and $\textbf{F}_{aux}^m$ are concatenated along the channel dimension. % along channel
   $\textbf{s}_{m}$ is learned by passing the concatenation to a three-layer MLP and a sigmoid function. 
   $\textbf{c}_{m}$ is learned by first taking global max pooling and average pooling to the concatenation and then passing to a three-layer MLP and sigmoid.} 
   \label{fig4_module}
\end{figure}

In each module, as illustrated in~\cref{fig4_module}, we learn a shared spatial coefficient matrix $\textbf{S}_m \in \mathbb{R}^{h\times w}$ and a shared channel coefficient vector $\textbf{c}_m \in~\mathbb{R}^c$ from the input feature pair $\textbf{F}_{vis}^m$ and $\textbf{F}_{aux}^m$ to model the dependency between the visible and auxiliary modalities at spatial and channel dimensions. 
% to model the spatial and channel dependency between the visible and auxiliary modalities. 
%
Then, to facilitate the information interaction between the two modalities, % 0129, interaction in ablation
$\textbf{F}_{vis}^m$ and $\textbf{F}_{aux}^m$ are updated as:
% \begin{equation}
%   {\textbf{F}^m_{vis}}' = \textbf{F}_{vis}^m + \textbf{S}_m \times  \textbf{F}_{aux}^m  + \textbf{c}_m \otimes \textbf{F}_{aux}^m,
%   \label{eq_f'vis}
% \end{equation}
% %
% \begin{equation}
%   {\textbf{F}^m_{aux}}' = \textbf{F}_{aux}^m + \textbf{S}_m \times \textbf{F}_{vis}^m + \textbf{c}_m \otimes \textbf{F}_{vis}^m,
%   \label{eq_f'aux}
% \end{equation}   
\begin{equation}
  {\textbf{F}^m_{vis}}' = \textbf{F}_{vis}^m + \textbf{S}_m \ast  \textbf{F}_{aux}^m  + \textbf{c}_m \circledast \textbf{F}_{aux}^m,
  \label{eq_f'vis}
\end{equation}
\begin{equation}
  {\textbf{F}^m_{aux}}' = \textbf{F}_{aux}^m + \textbf{S}_m \ast \textbf{F}_{vis}^m + \textbf{c}_m \circledast \textbf{F}_{vis}^m,
  \label{eq_f'aux}
\end{equation}                        
where $\ast$ and $\circledast$ denote spatial and channel-wise multiplication, respectively.
${\textbf{F}^m_{vis}}' \in \mathbb{R}^{h\times w\times c}$ and ${\textbf{F}^m_{aux}}' \in \mathbb{R}^{h\times w\times c}$ are then fed to the next stage in the encoders. 
Additionally,   
a fusion feature $\textbf{F}^m_{fus} \in \mathbb{R}^{h\times w\times c}$ is produced by fusing ${\textbf{F}^m_{vis}}'$ and ${\textbf{F}^m_{aux}}'$ via a $1\times 1$ convolution. % along the channel dimension.

The decoder predicts a segmentation mask ${\rm M} \in~\mathbb{R}^{H\times W}$ based on fusion features from the four modules.  % M和L没粗体
During training, the prediction of ${\rm M}$ is supervised by a ground-truth label ${\rm L} \in~\mathbb{R}^{H\times W}$ via a cross-entropy loss $\mathcal{L}_{ce} ({\rm M}, {\rm L})$.

Two projectors, following the encoders,
map final features $\textbf{F}_{vis}^4$ and $\textbf{F}_{aux}^4$ to representations $\textbf{R}_{vis} \in \mathbb{R}^{h\times w\times d}$ and $\textbf{R}_{aux} \in~\mathbb{R}^{h\times w\times d}$, respectively, which are utilized to generate embeddings in our supervised multi-modal contrastive learning approach. 
Detailed structure settings of the intermediate modules and the projectors are provided in~\cref{subsec:implementation}.

%------------------------------------------------------------------------

\subsection{Supervised Multi-Modal Contrastive Learning} 

Multi-modal dark scene understanding is challenging because the visible and auxiliary modalities both provide limited semantics.
We address this issue by introducing a supervised multi-modal contrastive learning approach~(\cref{fig2_mmcl}) to encourage the encoder feature spaces learned from the two modalities to be semantic-discriminative.

\textbf{Embedding Generation.}
To a set of visible-auxiliary representation pairs  
$\{\textbf{R}_{vis}^b, \textbf{R}_{aux}^b \in \mathbb{R}^{h\times w\times d} \}_{b=1}^B$
learned from a training batch~$B$ of input image pairs % ,
and a set of corresponding labels 
$\{\widetilde{\rm L}^b \in~\mathbb{R}^{h\times w} \}_{b=1}^B$
generated by downscaling the ground-truth labels, 
we sample a visible embedding set 
$\mathcal{V} = \{ \textbf{\textit{v}}_i \in \mathbb{R}^d : \textbf{\textit{v}}_i \to \widetilde{\rm L}_{\textbf{\textit{v}}_i}  \}$
% $\mathcal{V} = \{ \textbf{\textit{v}}_i \in \mathbb{R}^d, \widetilde{\rm L}_{\textbf{\textit{v}}_i}  \}$ 
and an auxiliary embedding set 
$\mathcal{A}  = \{ \textbf{\textit{a}}_j \in \mathbb{R}^d : \textbf{\textit{a}}_j \to \widetilde{\rm L}_{\textbf{\textit{a}}_j} \}$
from the representations.   
Taking visible embedding $\textbf{\textit{v}}_i$ as an example, 
$i$ denotes that it is sampled at the $i$-th spatial position in a visible representation, 
and $\widetilde{\rm L}_{\textbf{\textit{v}}_i}$ is its class label, 
which is obtained at the $i$-th position from the corresponding label
% and is utilized to align with its positive and negative samples. 
and is utilized to measure its class correlation with other embeddings. 
In both modalities,
% we randomly sample $n$ embeddings from each class present in the batch, % per
% equally divided among \textcolor{blue}{instances} of the class,
we randomly sample $n$ embeddings per instance from each class present in the batch,  % 句子正确? 另确认代码
and set $n$ as the number of pixels from the class with the least occurrences, 
following the protocol in~\cite{eccv22_mscs}. 
This setting maintains a balance for embeddings from each present class.
Then, 
% embeddings in $\mathcal{V}$ and $\mathcal{A}$ are cast to a shared space to perform contrast under the supervision of their class correlations: 
$\mathcal{V}$ and $\mathcal{A}$ are cast to a shared space to perform contrast under the supervision of the class correlations: 
Embeddings with the same label (or different labels) are same-class (or different-class) and are aligned as positive (or negative) samples.

\textbf{Cross-Modal Contrast.} 
The cross-modal contrast is to shape the visible and auxiliary feature spaces by considering 
% the correspondence of cross-modal contextual and geometry cues. 
the cross-modal context-geometry correspondence. 
To this end, we encourage embeddings from one modality to be closer to the same-class embeddings from the other modality % (positive)
and push apart different-class ones from across the two modalities % (negative), embeddings
by minimizing a cross-modal contrastive loss: 
\begin{equation}
  \mathcal{L}_{cm}(\mathcal{V}, \mathcal{A}) =
  \frac{1}{\mid \mathcal{V} \mid}
  \sum_{\textbf{\textit{v}}_i \in \mathcal{V}} 
  \frac{1}{\mid \mathcal{P}_{\textbf{\textit{v}}_i } \mid}
  \sum_{\textbf{\textit{a}}^+ \in \mathcal{P}_{\textbf{\textit{v}}_i }}
  \mathcal{L}_{\rm NCE}(\textbf{\textit{v}}_i , \textbf{\textit{a}}^+ ),
  \label{eq_cmterm}
\end{equation}
where
\begin{equation}   
\begin{split}
\begin{aligned}
  & \mathcal{L}_{\rm NCE}(\textbf{\textit{v}}_i , \textbf{\textit{a}}^+ ) = \\
  & - {\rm log} 
  \frac{{\rm exp}(\textbf{\textit{v}}_i \cdot\textbf{\textit{a}}^+ /\tau )}
  {{\rm exp}(\textbf{\textit{v}}_i \cdot\textbf{\textit{a}}^+ /\tau ) +
  \sum_{\textbf{\textit{a}}^- \in\mathcal{N}_{\textbf{\textit{v}}_i }} {\rm exp}(\textbf{\textit{v}}_i \cdot\textbf{\textit{a}}^- /\tau )}
  \label{eq_cminfo}
\end{aligned}
\end{split}
\end{equation}  
is the InfoNCE loss~\cite{arxiv18_infonce}. % aistats10_infonce / iccv21 cross-image
The symbol $\cdot$ denotes the dot product. $\tau$ is a temperature hyperparameter. 
$\mathcal{P}_{\textbf{\textit{v}}_i } = \{ \textbf{\textit{a}}_j \in \mathcal{A} : j \ne i, \widetilde{\rm L}_{\textbf{\textit{a}}_j} = 
\widetilde{\rm L}_{\textbf{\textit{v}}_i} \}$ 
and % i\mid j
$\mathcal{N}_{\textbf{\textit{v}}_i } = \{ \textbf{\textit{a}}_j \in \mathcal{A} : j \ne i, \widetilde{\rm L}_{\textbf{\textit{a}}_j} \ne \widetilde{\rm L}_{\textbf{\textit{v}}_i} \}$ 
% are respectively the sets of positive and negative auxiliary embeddings, \textit{i.e.,} samples, for visible embedding~$\textbf{\textit{v}}_i$. 
are respectively the sets of same-class and different-class auxiliary embeddings, \textit{i.e.,} positive and negative samples, for visible embedding~$\textbf{\textit{v}}_i$.
% 0114, 需要说明位置不对应, 看get masks. 另比如n, 位置相同就同类, 就不是n了. p的i|j 看p8...
%       另, 前面需要位置才能得到类信息
%
Note that, since the positive and negative relation among the embeddings is bidirectional, the cross-modal contrast has only one loss term. % ~\cite{iccv21_crossclr,cvpr22_crosspoint} 顺序, 或哪个? mscs

\begin{figure*}
  \centering
  \begin{subfigure}{0.162\linewidth}
    \includegraphics[width=1\linewidth]{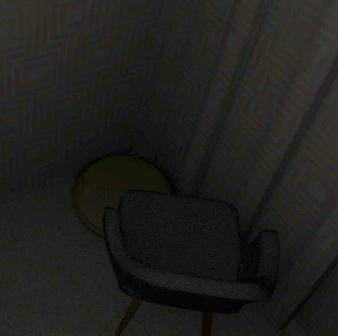}
    \includegraphics[width=1\linewidth]{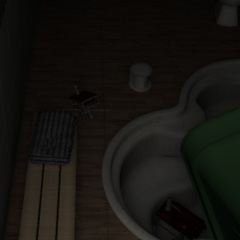}
    \caption{RGB}
  \end{subfigure}
  \begin{subfigure}{0.162\linewidth}
    \includegraphics[width=1\linewidth]{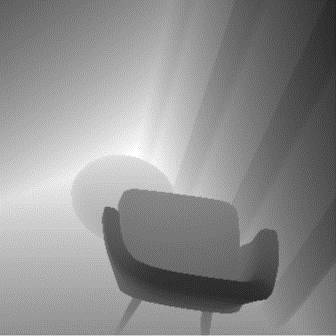}
    \includegraphics[width=1\linewidth]{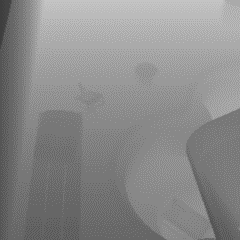}
    \caption{Depth}
  \end{subfigure}
  \begin{subfigure}{0.162\linewidth}
    \includegraphics[width=1\linewidth]{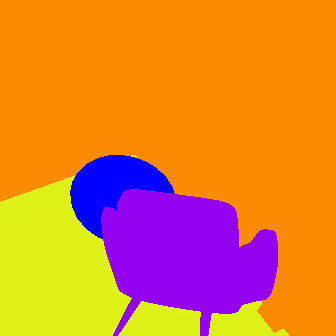}
    \includegraphics[width=1\linewidth]{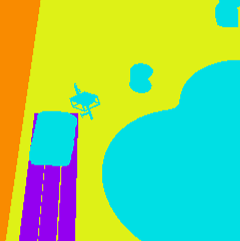}
    \caption{Ground Truth}
  \end{subfigure}
  \begin{subfigure}{0.162\linewidth}
    \includegraphics[width=1\linewidth]{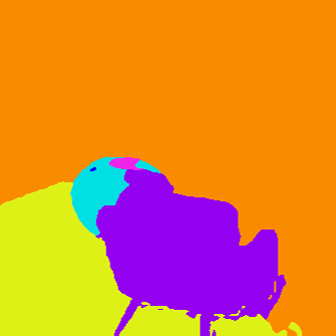}
    \includegraphics[width=1\linewidth]{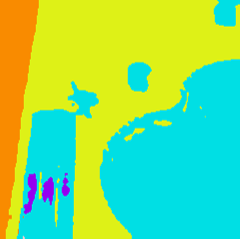}
    \caption{TokenFusion~\cite{cvpr22_token}}
  \end{subfigure}
  \begin{subfigure}{0.162\linewidth}
    \includegraphics[width=1\linewidth]{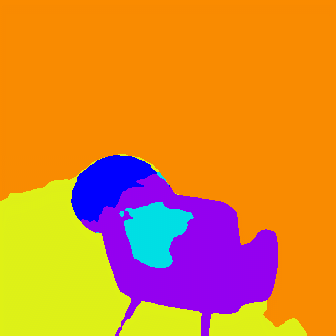}
    \includegraphics[width=1\linewidth]{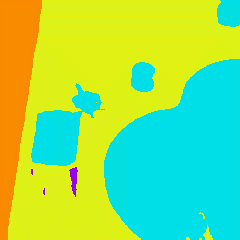}
    \caption{CMX~\cite{2022_cmx}}
  \end{subfigure}
  \begin{subfigure}{0.162\linewidth}
    \includegraphics[width=1\linewidth]{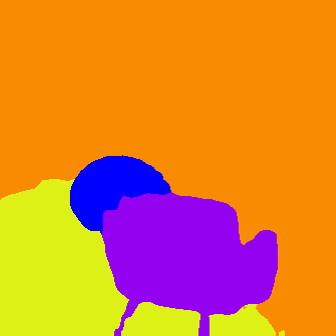}
    \includegraphics[width=1\linewidth]{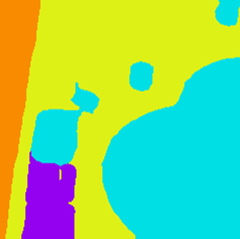}
    \caption{Ours}
  \end{subfigure}
  \caption{
  Low-light indoor scene segmentation from RGB-depth data. 
  Visual comparisons between Base and Ours are provided in~\cref{subsec:ablation} 
  and the supplementary material.} 
  \label{fig5_llrgbd}
\end{figure*}

\textbf{Intra-Modal Contrast.}
The intra-modal contrast shapes the encoder feature spaces of the two modalities by regularizing embeddings within each modality separately.
Within the visible modality, same-class or different-class embeddings are pulled closer or pushed apart 
by minimizing an intra-modal contrastive loss:
\begin{equation}
  \mathcal{L}_{vis}(\mathcal{V}) =
  \frac{1}{\mid \mathcal{V} \mid}
  \sum_{\textbf{\textit{v}}_i \in \mathcal{V}} 
  \frac{1}{\mid \mathcal{P}'_{\textbf{\textit{v}}_i } \mid}
  \sum_{\textbf{\textit{v}}^+ \in \mathcal{P}'_{\textbf{\textit{v}}_i }}
  \mathcal{L}_{\rm NCE}(\textbf{\textit{v}}_i , \textbf{\textit{v}}^+),
  \label{eq_visterm}
\end{equation}
where
\begin{equation}   
\begin{split}
\begin{aligned}
  & \mathcal{L}_{\rm NCE}(\textbf{\textit{v}}_i , \textbf{\textit{v}}^+) = \\
  & - {\rm log} 
  \frac{{\rm exp}(\textbf{\textit{v}}_i \cdot\textbf{\textit{v}}^+ /\tau )}
  {{\rm exp}(\textbf{\textit{v}}_i \cdot\textbf{\textit{v}}^+ /\tau ) +
  \sum_{\textbf{\textit{v}}^- \in\mathcal{N}'_{\textbf{\textit{v}}_i }} {\rm exp}(\textbf{\textit{v}}_i \cdot\textbf{\textit{v}}^- /\tau )}.
  \label{eq_visinfo}
\end{aligned}
\end{split}
\end{equation}  
%
% 这里p n sets 看p6
$\mathcal{P}'_{\textbf{\textit{v}}_i } = \{ \textbf{\textit{v}}_p  \in \mathcal{V} \mid p \ne i, \widetilde{\rm L}_{\textbf{\textit{v}}_p} = 
\widetilde{\rm L}_{\textbf{\textit{v}}_i} \}$ 
and 
$\mathcal{N}'_{\textbf{\textit{v}}_i } = \{ \textbf{\textit{v}}_p \in \mathcal{V} \mid p \ne i, \widetilde{\rm L}_{\textbf{\textit{v}}_p} \ne \widetilde{\rm L}_{\textbf{\textit{v}}_i} \}$ are respectively the intra-modal sets of positive and negative samples for $\textbf{\textit{v}}_i$. 
The contrastive loss for the auxiliary modality, \textit{i.e.},~$\mathcal{L}_{aux}(\mathcal{A})$, is similar to~\cref{eq_visterm}.
For $\textbf{\textit{a}}_j \in \mathcal{A}$, the positive and negative sample sets are
$\mathcal{P}'_{\textbf{\textit{a}}_j } = \{ \textbf{\textit{a}}_q \in \mathcal{A} \mid q \ne j, \widetilde{\rm L}_{\textbf{\textit{a}}_q} = 
\widetilde{\rm L}_{\textbf{\textit{a}}_j} \}$ 
and 
$\mathcal{N}'_{\textbf{\textit{a}}_j } = \{ \textbf{\textit{a}}_q \in \mathcal{A} \mid q \ne j, \widetilde{\rm L}_{\textbf{\textit{a}}_q} \ne \widetilde{\rm L}_{\textbf{\textit{a}}_j} \}$,
respectively.

Combining the cross-modal contrastive loss and the intra-modal contrastive losses, 
our full training objective is: 
\begin{equation}   
  \mathcal{L} = \mathcal{L}_{ce} + \lambda_{cm}\mathcal{L}_{cm} + \lambda_{vis}\mathcal{L}_{vis} + \lambda_{aux}\mathcal{L}_{aux}.
  \label{eq_full}
\end{equation}  
% 
% 3.2开头, 摘要, 引言 贡献, 2.3末尾
Experiments in~\cref{sec:experiments} show that our approach can effectively enhance dark scene understanding by shaping semantic-discriminative feature spaces.

%------------------------------------------------------------------------
%------------------------------------------------------------------------

\section{Experiments} 
\label{sec:experiments} 

\subsection{Implementation Details}  
\label{subsec:implementation}

\textbf{Network Structure.}
We employ three different backbones, including ResNet-101, SegFomrer-B2~\cite{nips_21segformer}, and SegNext-B~\cite{nips22_segnext}, to build our segmentation network. 
The channel setting to the four encoder stages is $[c_1, c_2, c_3, c_4] = [64, 128, 320, 512]$. % 256
Taking the first of the four intermediate modules as an example, 
the input and output channel setting of the MLP layers for spatial and channel coefficient learning is listed in~\cref{table1_mlp}, 
and the input and output channels of the $1\times 1$ convolution used for feature fusion are set as $[2c_1, c_1]$. 
The two projectors each consist of a two-layer MLP and a linear mapping with $d=256$, 
where the input and output channels of the MLP layers are equal to $c_4$. 

\begin{table}[h]
\begin{center}
\begin{tabular}{|c|ccc|}
\hline
        % & MLP$_1$         & MLP$_2$          & MLP$_3$     \\ \hline \hline
        & Layer$_1$       & Layer$_2$        & Layer$_3$     \\ \hline \hline  % 0307
Spatial & $[2c_1, 2c_1]$  &  $[2c_1, 2c_1]$  & $[2c_1, 1]$ \\ \hline
Channel & $[4c_1, 4c_1]$  &  $[4c_1, 4c_1]$  & $[4c_1, c_1]$ \\
\hline
\end{tabular}
\end{center}
\vspace{-0.25cm}
\caption{
Channel setting, $[in, out]$, of the MLP layers for spatial and channel coefficient learning in the first intermediate module.} 
\label{table1_mlp}
\end{table}

\textbf{Contrastive Losses.} 
The weights $\lambda_{cm}$, $\lambda_{vis}$, and $\lambda_{aux}$ in~\cref{eq_full} are set as 0.2 in experiments for low-light indoor scene segmentation, %including the ablation study, 
and are set as 0.05 in experiments for nighttime outdoor scene and normal-light scene segmentation. % segmentation
The temperature $\tau$ in $\mathcal{L}_{cm}$, $\mathcal{L}_{vis}$, and $\mathcal{L}_{aux}$ is set as 0.1 in all experiments.
Ablations on $\lambda_{cm}$, $\lambda_{vis}$, $\lambda_{aux}$, and $\tau$ are provided in the supplementary material.

\begin{figure*}
  \centering
  \begin{subfigure}{0.162\linewidth}
    \includegraphics[width=1\linewidth]{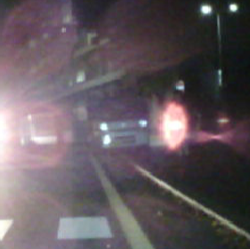}
    \includegraphics[width=1\linewidth]{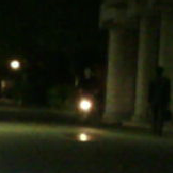}
    \caption{RGB}
  \end{subfigure}
  \begin{subfigure}{0.162\linewidth}
    \includegraphics[width=1\linewidth]{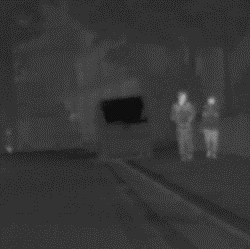}
    \includegraphics[width=1\linewidth]{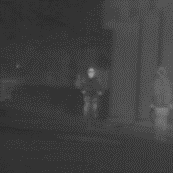}
    \caption{Thermal}
  \end{subfigure}
  \begin{subfigure}{0.162\linewidth}
    \includegraphics[width=1\linewidth]{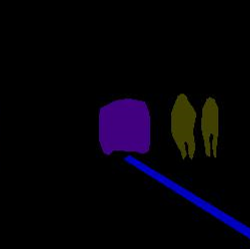}
    \includegraphics[width=1\linewidth]{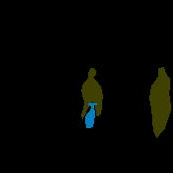}
    \caption{Ground Truth}
  \end{subfigure}
  \begin{subfigure}{0.162\linewidth}
    \includegraphics[width=1\linewidth]{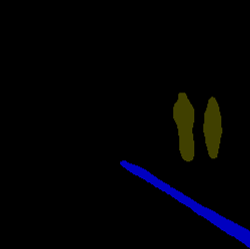}
    \includegraphics[width=1\linewidth]{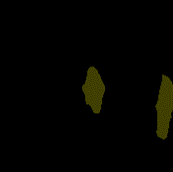}
    \caption{TokenFusion~\cite{cvpr22_token}}
  \end{subfigure}
  \begin{subfigure}{0.162\linewidth}
    \includegraphics[width=1\linewidth]{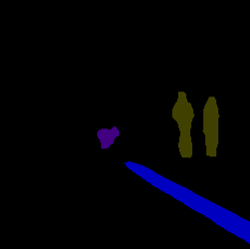}
    \includegraphics[width=1\linewidth]{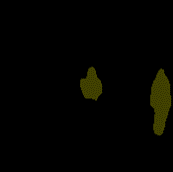}
    \caption{CMX~\cite{2022_cmx}}
  \end{subfigure}
  \begin{subfigure}{0.162\linewidth}
    \includegraphics[width=1\linewidth]{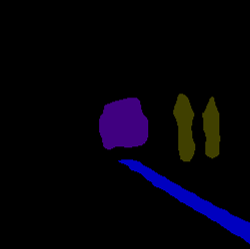}
    \includegraphics[width=1\linewidth]{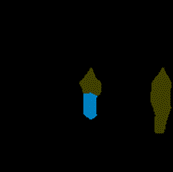}
    \caption{Ours}
  \end{subfigure}
  \caption{Nighttime outdoor scene segmentation from RGB-thermal data.} 
  \label{fig6_mfnet}
\end{figure*}

% Backbone}

\begin{table}[t] 
\begin{center}
\begin{tabular}{|c|c|c|}
\hline
Method        & Backbone & mIoU (\%) \\ \hline \hline
 % \multicolumn{3}{|c|}{CNN-based} \\ \hline 
SA-Gate$^{\dag}$~\cite{eccv20_sagate}    & ResNet-101 & 61.79  \\ \hline 
ShapeConv$^{\dag\S}$~\cite{iccv21_shape} & ResNeXt-101 & 63.26 \\ \hline
CEN$^{\dag}$~\cite{pami22_cen}           & ResNet-101 & 62.15     \\ \hline 
 % \multicolumn{3}{|c|}{Transformer-based} \\ \hline  
TokenFusion$^{\dag}$~\cite{cvpr22_token} & SegFormer-B2 & 64.75       \\ \hline
CMX$^{\dag}$~\cite{2022_cmx}             & SegFormer-B2 & 66.52 \\ \hline \hline
\multirow{3}{*}{Base {\small (\textit{w/o} SMMCL)}} & ResNet-101   & 62.73  \\  \cline{2-3}
                                                   & SegFormer-B2 & 65.69  \\  \cline{2-3}
                                                   & SegNeXt-B    & 66.02  \\ \hline \hline 
\multirow{3}{*}{Ours {\small (\textit{w} SMMCL)}}   & ResNet-101   & 64.40  \\ \cline{2-3}
                                                   & SegFormer-B2 & 67.77  \\ \cline{2-3}
                                                   & SegNeXt-B    & \textbf{68.76} \\
\hline
\end{tabular}
\end{center}
\vspace{-0.25cm}
\caption{Low-light indoor scene segmentation from RGB-depth data. 
Single-scale results are reported by default. 
$^{\dag}$Our implementation. 
$^{\S}$Multi-scale results.  
The best result is shown in \textbf{bold}. 
% The second best result is {\ul underlined}.
}  
\label{table2_llrgbd}
\end{table}

% see config.py...... and next, sa, cmx 
\textbf{Training and Evaluation.}  
We implement our model with PyTorch on four Tesla V100 GPUs. 
During training, we minimize the objective in~\cref{eq_full}. 
The encoders are initialized with the ImageNet-1K pretrained weights. % ~\cite{cvpr09_imagenet}
We employ AdamW~\cite{iclr19_adamw} optimizer.  % decay 0.01
The initial learning rate is $6e^{-5}$ and decays following the poly policy. % power 0.9 by default, warm up 10   
We use basic augmentation techniques, % common, standard
including random horizontal flipping and random scaling from 0.5 to 1.75.  % no random cropping 
The batch size is 16. 
We adopt the above training setting in all experiments.
In low-light, nighttime, and normal-light scene segmentation tasks, we train our model for 500, 300, and 600 epochs, respectively. 
During evaluation, we use mean Intersection over Union (mIoU) as the metric. 
We do not use any tricks, \textit{e.g.}, multi-scale inference, when evaluating our model.

%------------------------------------------------------------------------

\subsection{Validation on Low-Light Indoor Scenes} 
\label{subsec:llrgbd}

% llrgbd: synthetic, low-light indoor, rgb-depth, 13 
\textbf{Dataset and Comparison Methods.} 
We conduct the task of understanding low-light indoor scenes from RGB-depth data on the LLRGBD-synthetic dataset~\cite{2021_llrgbd}. 
LLRGBD-synthetic is a large-scale synthetic dataset with 13 semantic classes. 
To lower data redundancy, we randomly sample 1418 scenes from its training set for training, 
and sample 479 scenes from its validation set for evaluation. 
We compare our model with five state-of-the-art multi-modal image segmentation methods: CMX~\cite{2022_cmx}, TokenFusion~\cite{cvpr22_token}, CEN~\cite{pami22_cen}, ShapeConv~\cite{iccv21_shape}, and SA-Gate~\cite{eccv20_sagate}.

\textbf{Results.} 
Quantitative comparison results are reported in~\cref{table2_llrgbd}. 
As can be observed, in low-light indoor scenes,
our model trained with the proposed supervised multi-modal contrastive learning approach achieves segmentation accuracy of $68.76\% / 67.77\% / 64.40\%$, and results in a $2.74\% / 2.08\% / 1.67\%$ improvement over the baseline\footnote{The baseline, \textit{i.e.,} Base~{\small(\textit{w/o} SMMCL)}, employs the same network structure in~\cref{fig3_model},
but is trained with only a cross-entropy loss.}. 
Besides,
in comparison with the five state-of-the-art methods, our model achieves the highest accuracy, 
and outperforms them by a large margin.
\Cref{fig5_llrgbd} shows segmentation masks predicted by CMX, TokenFusion, and our best model.
Due to a lack of consideration for the class correlations, 
CMX and TokenFusion tend to predict incorrect class information in the scenes, 
where the poor visibility in the RGB modality and the lack of contextual semantics of the depth modality cause low class discrimination. 
By contrast, our model can segment the scenes with much higher accuracy. 
This is because our supervised multi-modal contrastive learning approach fully considers the correlations among semantic classes,  % 上面说了别人缺乏对correlations的考虑
and can effectively enhance multi-modal dark scene understanding by shaping semantic-discriminative feature spaces.
We provide comprehensive ablation supports in~\cref{subsec:ablation}.

\begin{table}[t] 
\begin{center}
\begin{tabular}{|c|c|c|}
\hline
Method        & Backbone & mIoU (\%) \\ \hline \hline
 % \multicolumn{3}{|c|}{CNN-based} \\ \hline 
RTFNet~\cite{2019_rtfnet}     & ResNet-152 & 54.8     \\ \hline
GMNet~\cite{tip21_gmnet}      & ResNet-50 & 57.7     \\ \hline
ABMDRNet~\cite{cvpr21_abmdernet} & ResNet-50  & 55.5 \\ \hline  
LASNet~\cite{tcsvt22_lasnet}  & ResNet-152   & 58.7     \\ \hline 
 % \multicolumn{3}{|c|}{Transformer-based} \\ \hline  
TokenFusion$^{\dag}$~\cite{cvpr22_token}  & SegFormer-B2 & 58.7  \\ \hline  
CMX~\cite{2022_cmx}  & SegFormer-B2 & 57.8  \\ \hline \hline % order
\multirow{3}{*}{Base {\small (\textit{w/o} SMMCL)}} & ResNet-101   & 57.2  \\  \cline{2-3}
                                                   & SegFormer-B2 & 57.9  \\  \cline{2-3}
                                                   & SegNeXt-B    & 58.4   \\ \hline \hline 
\multirow{3}{*}{Ours {\small (\textit{w} SMMCL)}}   & ResNet-101   & 58.9  \\ \cline{2-3}
                                                   & SegFormer-B2 & 59.8  \\ \cline{2-3}
                                                   & SegNeXt-B    & \textbf{60.0} \\
\hline
\end{tabular}
\end{center}
\vspace{-0.25cm}
\caption{Nighttime outdoor scene segmentation on RGB-thermal data. 
Single-scale results are reported. % by default.
$^{\dag}$Our implementation. 
The best result is shown in \textbf{bold}. 
% The second best result is {\ul underlined}.
}  
\label{table3_mfnet}
\end{table}

%-------------------------------------------------------------------------

\begin{figure*}
  \centering
  \begin{subfigure}{0.162\linewidth}
    \includegraphics[width=1\linewidth]{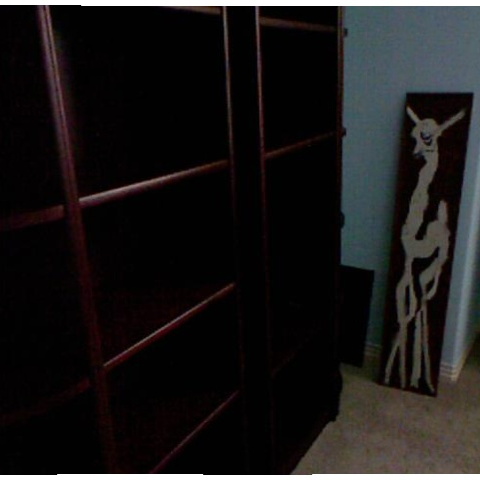}
    \includegraphics[width=1\linewidth]{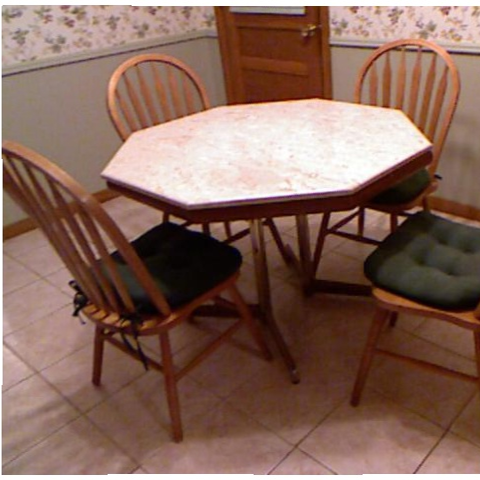}
    \caption{RGB}
  \end{subfigure}
  \begin{subfigure}{0.162\linewidth}
    \includegraphics[width=1\linewidth]{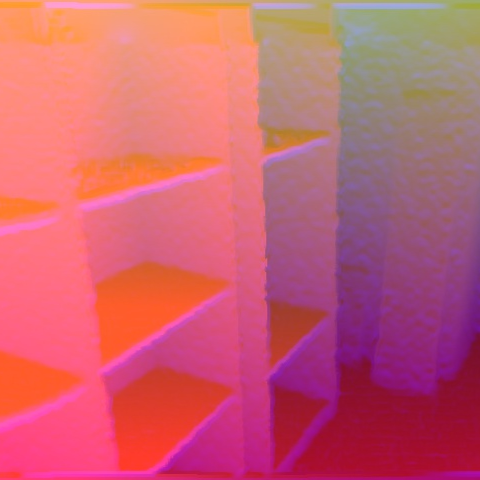}
    \includegraphics[width=1\linewidth]{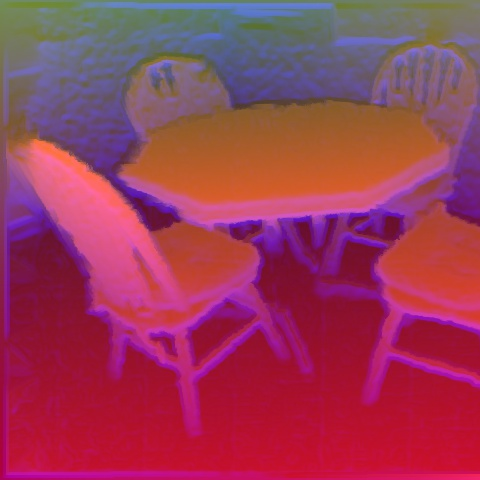}
    \caption{HHA}
  \end{subfigure}
  \begin{subfigure}{0.162\linewidth}
    \includegraphics[width=1\linewidth]{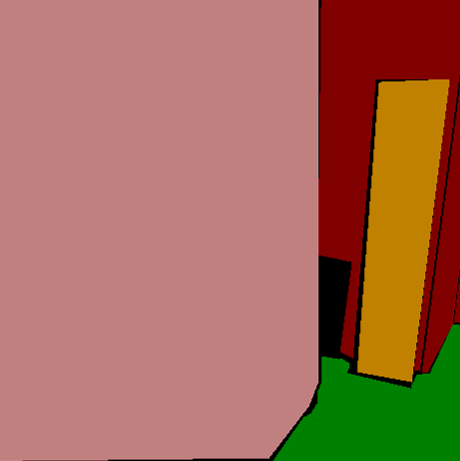}
    \includegraphics[width=1\linewidth]{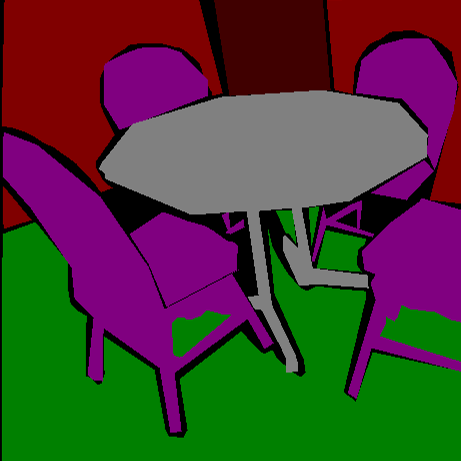}
    \caption{Ground Truth}
  \end{subfigure}
  \begin{subfigure}{0.162\linewidth}
    \includegraphics[width=1\linewidth]{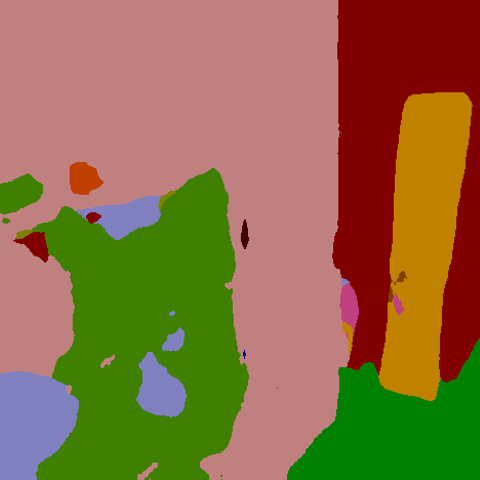}
    \includegraphics[width=1\linewidth]{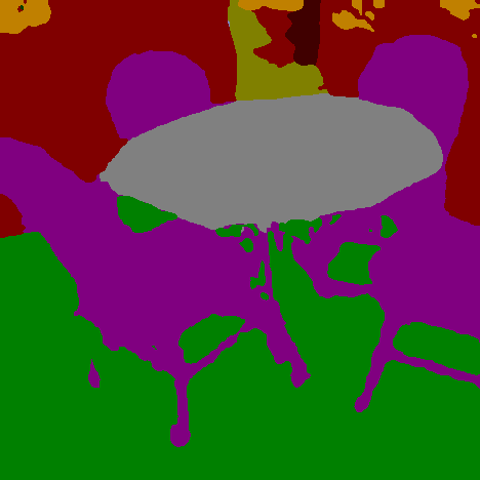}
    \caption{TokenFusion~\cite{cvpr22_token}}
  \end{subfigure}
  \begin{subfigure}{0.162\linewidth}
    \includegraphics[width=1\linewidth]{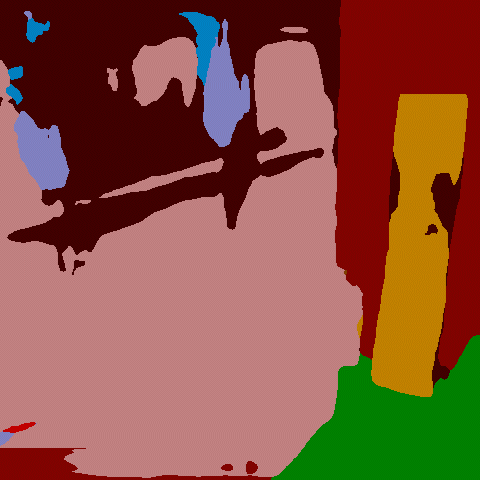}
    \includegraphics[width=1\linewidth]{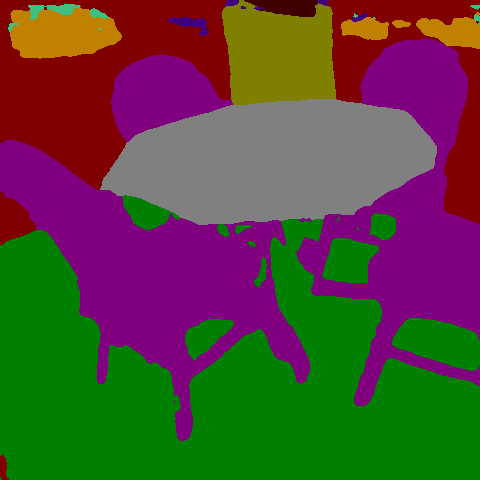}
    \caption{CMX~\cite{2022_cmx}}
  \end{subfigure}
  \begin{subfigure}{0.162\linewidth}
    \includegraphics[width=1\linewidth]{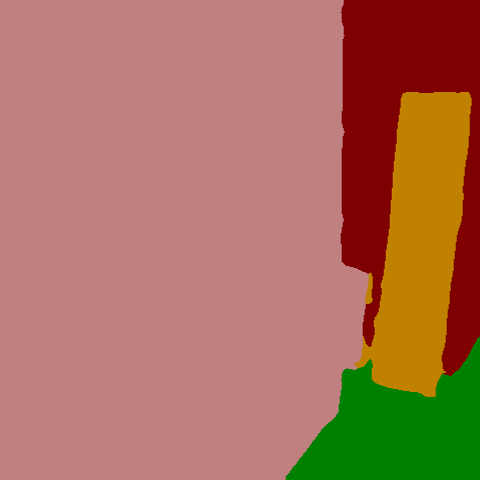}
    \includegraphics[width=1\linewidth]{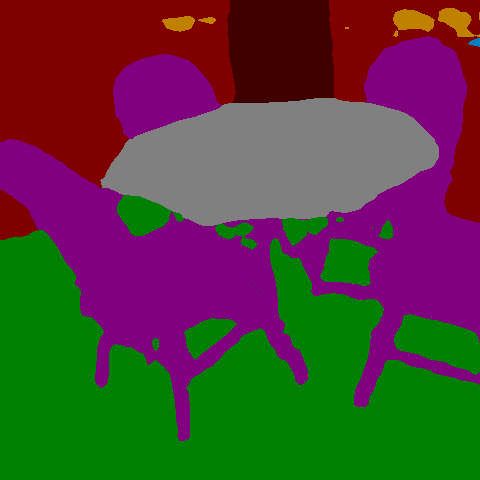}
    \caption{Ours}
  \end{subfigure}
  \caption{Normal-light scene segmentation from RGB-depth data. 
   Depth images are encoded to HHA maps~\cite{eccv14_hha} in this task. }
  \label{fig7_nyudv2}
\end{figure*}

\subsection{Validation on Nighttime Outdoor Scenes} 

% mfnet: real-world, nighttime outdoor, rgb-thermal, 9 
\textbf{Dataset and Comparison Methods.}
We further validate our method on real-world nighttime outdoor scenes using RGB-thermal data from the MFNet dataset~\cite{iros17_mfnet}.
MFNet provides 1569 outdoor scenes covering daytime and nighttime, with 9 semantic classes. 
We train our model with all 784 scenes in the training set and evaluate it on the 188 nighttime scenes in the test set. 
We compare our model with TokenFusion~\cite{cvpr22_token} and
five state-of-the-art RGB-thermal segmentation methods: 
CMX~\cite{2022_cmx}, ABMDRNet~\cite{cvpr21_abmdernet}, LASNet~\cite{tcsvt22_lasnet}, GMNet~\cite{tip21_gmnet}, and RTFNet~\cite{2019_rtfnet}.

\textbf{Results.}
\Cref{table3_mfnet} reports the quantitative comparison results. 
In nighttime outdoor scenes,
our model with supervised multi-modal contrastive learning achieves the highest accuracy of $60.0\% / 59.8\% / 58.9\%$, and gains a $1.6\% / 1.9\% / 1.7\%$ improvement over the baseline. 
Moreover, our best model outperforms the second best method, TokenFusion, by $1.3\%$. 
\Cref{fig6_mfnet} qualitatively compares our best model with TokenFusion and CMX. 
As shown, they fail to segment the car in the first scene and the riding man in the second scene, since the RGB and thermal modalities provide limited semantic cues for the two objects. 
In contrast,
our model predicts more accurate segmentation masks for these two difficult cases. 
This is due to our supervised multi-modal contrastive learning approach enables our model to better understand scenes from multi-modal images with limited semantics. 
We present more qualitative comparisons in the supplementary material.

%------------------------------------------------------------------------

\subsection{Generalizability on Normal-Light Scenes} 

% nyuv2: real-world, normal-light outdoor, rgb-hha, 40 
\textbf{Dataset and Comparison Methods.}
We validate the generalization capability of our approach on real-world normal-light scenes using RGB-depth data from the NYUDv2 dataset~\cite{eccv12_nyu}.
NYUDv2 dataset provides 1449 indoor scenes with 40 semantic classes, in which 795 scenes are for training and 654 scenes are for evaluation. 
We compare our model with five state-of-the-art RGB-depth segmentation methods: CMX~\cite{2022_cmx}, TokenFusion~\cite{cvpr22_token}, CEN~\cite{pami22_cen}, ShapeConv~\cite{iccv21_shape}, and SA-Gate~\cite{eccv20_sagate}.

\textbf{Results.} 
Quantitative comparisons are shown in~\cref{table4_nyudv2}.
% Our approach brings a $1.1\% / 1.4\% / 0.9\%$ improvement over the baseline, and allows our model to achieve the best accuracy, $55.8\%$, which is \textcolor{cyan}{$1.0\%$} higher than the second best method, TokenFusion.
Our approach brings a $1.1\% / 1.4\% / 0.9\%$ improvement over the baseline. Besides, our best model outperforms the second best method, TokenFusion~(SegFormer-B3), by $1.0\%$. 
\Cref{fig7_nyudv2} qualitatively compares our best model with TokenFusion~(SegFormer-B3) and CMX. 
Our model shows superior generalizability in normal-light scenes. 
While the other two methods fail to predict correct class information for the dark areas in the first scene and the door and wallpaper in the second scene, 
our model achieves predictions closer to the ground truth. 
This is because 
our approach enables our model to capture the classes similarities and differences more accurately and understand scenes from multi-modal images more effectively. 
We demonstrate this point via visual comparisons between Base and Ours in~\cref{subsec:ablation}.

%------------------------------------------------------------------------

\begin{table}[t] 
\begin{center}
\begin{tabular}{|c|c|c|}
\hline
Method        & Backbone & mIoU (\%) \\ \hline \hline
 % \multicolumn{3}{|c|}{CNN-based} \\ \hline 
SA-Gate$^{\S}$~\cite{eccv20_sagate}    & ResNet-101 & 52.4  \\ \hline % 101: see github和supp
% single-scale: you will expect ~51.4% mIoU in SA-Gate.nyu and ~51.5% mIoU in SA-Gate.nyu.432. single scale inference with no flip.
ShapeConv$^{\S}$~\cite{iccv21_shape} & ResNeXt-101 & 51.3 \\ \hline
CEN~\cite{pami22_cen}           & ResNet-101 & 51.1     \\ \hline 
 % \multicolumn{3}{|c|}{Transformer-based} \\ \hline  
TokenFusion~\cite{cvpr22_token} & SegFormer-B2 & 53.3 \\ \hline % b2免了b3被当typo, 且体现b3更强
TokenFusion~\cite{cvpr22_token} & SegFormer-B3 & 54.8 \\ \hline % 放b2后, 55.1和54.1, 54.7和54.8
CMX~\cite{2022_cmx}             & SegFormer-B2 & 54.1 \\ \hline \hline  % supp 
\multirow{3}{*}{Base {\small (\textit{w/o} SMMCL)}} & ResNet-101   & 52.5  \\  \cline{2-3}
                                                   & SegFormer-B2 & 53.7  \\  \cline{2-3}
                                                   & SegNeXt-B    & 54.7   \\ \hline \hline 
\multirow{3}{*}{Ours {\small (\textit{w} SMMCL)}}   & ResNet-101   & 53.4  \\ \cline{2-3}
                                                   & SegFormer-B2 & 55.1   \\ \cline{2-3}
                                                   & SegNeXt-B    & \textbf{55.8} \\
                                                   
\hline
\end{tabular}
\end{center}
\vspace{-0.25cm}
\caption{Normal-light scene segmentation from RGB-depth data. 
Single-scale results are reported by default. 
$^{\S}$Multi-scale results. 
% $^{\dag}$Our implementation. 
The best result is shown in \textbf{bold}. 
% The second best result is {\ul underlined}.
}  
\label{table4_nyudv2}
\end{table}

\begin{figure}
  \centering
  \begin{subfigure}{0.33\linewidth}  
    \includegraphics[width=1\linewidth]{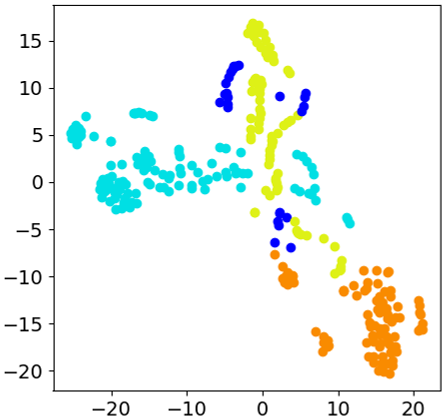}
    \caption{Visible~(Base)}
  \end{subfigure}
  {~~~~~~}  % 230815
  \begin{subfigure}{0.33\linewidth}
    \includegraphics[width=1\linewidth]{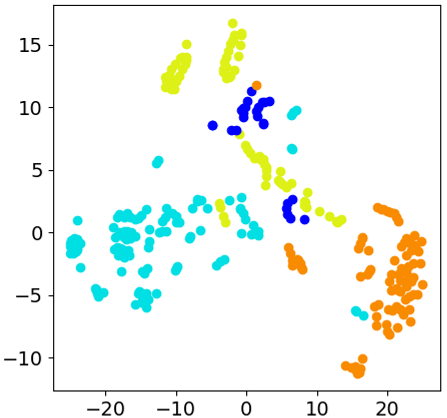}
    \caption{Auxiliary~(Base)}
  \end{subfigure}
  \begin{subfigure}{0.33\linewidth}
    \includegraphics[width=1\linewidth]{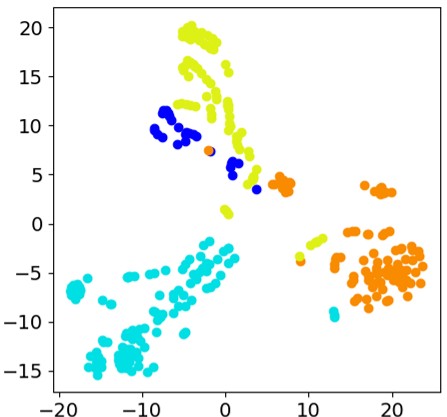}
    \caption{Visible~(Ours)}
  \end{subfigure}
  {~~~~~~}
  \begin{subfigure}{0.33\linewidth}
    \includegraphics[width=1\linewidth]{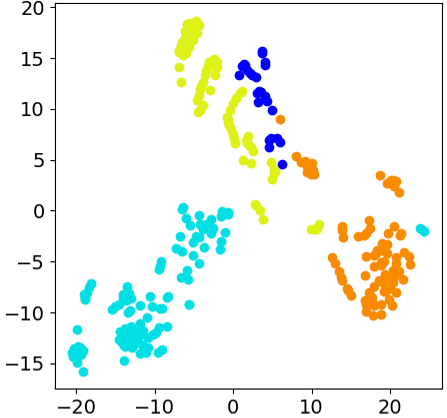}
    \caption{Auxiliary~(Ours)}
  \end{subfigure}
   \caption{
   TSNE visualization~\cite{2008_tsne} for final encoder features from Base~{\small(\textit{w/o} SMMCL)} and Ours~{\small(\textit{w}~SMMCL)} on a low-light scene in LLRGBD-synthetic. 
   Each color corresponds to a semantic class.  
   }
   \vspace{-0.35cm}
   \label{fig8_tsne}
\end{figure}

\begin{table}[t]
\begin{center}
\begin{tabular}{|c|c|c|c|c|}
\hline
    & Model$_1$    & Model$_2$   & Model$_3$  & Model$_4$  \\ \hline \hline 
Cross-Modal & {\small \ding{55}} &  \textbf{\checkmark} & {\small \ding{55}} & \textbf{\checkmark} \\ \hline    
% Cross Contrast & {\small \ding{55}} &  \textbf{\checkmark} & {\small \ding{55}} & \textbf{\checkmark} \\ \hline   
Intra-Modal & {\small \ding{55}} &  {\small \ding{55}}  & \textbf{\checkmark} & \textbf{\checkmark} \\ \hline    
% Intra-Modal & {\small \ding{55}} &  {\small \ding{55}}  & \textbf{\checkmark} & \textbf{\checkmark} \\ \hline  
mIoU (\%) & 66.02 & 68.62 & 68.52 & 68.76 \\ 
\hline
\end{tabular}
\end{center}
\vspace{-0.25cm}
\caption{
Effectiveness study of our supervised multi-modal contrastive learning approach on low-light indoor scenes.
}
\label{table5_contrasts}
\end{table}

\subsection{Ablation Study}
\label{subsec:ablation}

We thoroughly analyze our supervised multi-modal contrastive learning approach in this section\footnote{We conduct ablations using our model with the SegNeXt-B backbone.}.
Other ablations are provided in the supplementary material.  

%------------------------------------------------------------------------

% 要呼应4.2 4.3 4.4

\textbf{Basic Ablations.} 
\Cref{table5_contrasts} studies the effectiveness of our approach on low-light indoor scenes.  
In a comparison of Model$_1$, \textit{i.e.}, Base~{\small(\textit{w/o} SMMCL)}, which is trained with only a cross-entropy loss, 
and Model$_2$, which is trained by adding the cross-modal contrastive loss, 
Model$_2$ yields a $2.6\%$ improvement and accuracy of $68.62\%$. 
By adding the intra-modal contrastive losses, 
Model$_3$ produces accuracy of $68.52\%$. 
Further,
by jointly introducing cross-modal and intra-modal contrast, 
Model$_4$, \textit{i.e.},~Ours~{\small(\textit{w}~SMMCL)}, 
achieves the best accuracy, $68.76\%$.

\textbf{TSNE Visualization.} 
\Cref{fig8_tsne} visualizes the final encoder features, 
\textit{i.e.,} $\textbf{F}_{vis}^4$ and $\textbf{F}_{aux}^4$, 
learned by Base~{\small(\textit{w/o} SMMCL)} and Ours~{\small(\textit{w} SMMCL)}. 
%
% 要呼应4.2 4.3 4.4, 及引言 相关 贡献 方法等 
As shown in subfigures (a-b), features learned by Base~{\small(\textit{w/o}~SMMCL)} show low semantic discriminability, with points from different classes being in a mixed distribution.~By contrast, in features learned by Ours~{\small(\textit{w}~SMMCL)}, \textit{i.e.,}~subfigures (c-d), points belonging to the same class are closer and form clearer clusters.
This demonstrates that our approach 
effectively
encourages the feature spaces learned from multi-modal images with limited semantics to show higher semantic discriminability.

\begin{figure}
  \centering
  \begin{subfigure}{0.19\linewidth}
    \includegraphics[width=1\linewidth]{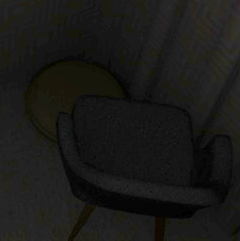}
    \includegraphics[width=1\linewidth]{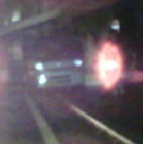}
    \includegraphics[width=1\linewidth]{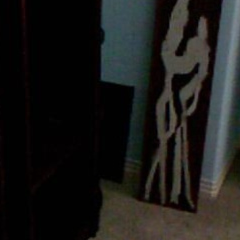}
    \caption{Visible}
  \end{subfigure}
  \begin{subfigure}{0.19\linewidth}
    \includegraphics[width=1\linewidth]{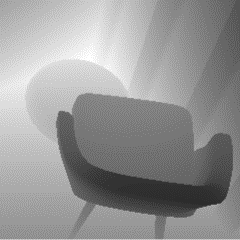}
    \includegraphics[width=1\linewidth]{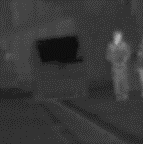}
    \includegraphics[width=1\linewidth]{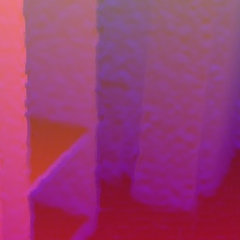}
    \caption{Auxiliary}
  \end{subfigure}
  \begin{subfigure}{0.19\linewidth}
    \includegraphics[width=1\linewidth]{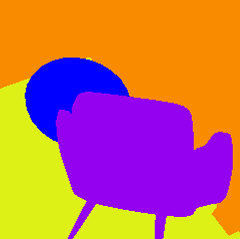}
    \includegraphics[width=1\linewidth]{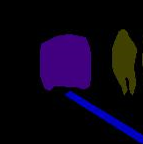}
    \includegraphics[width=1\linewidth]{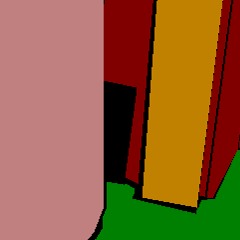}
    \caption{GT}
  \end{subfigure}
  \begin{subfigure}{0.19\linewidth}
    \includegraphics[width=1\linewidth]{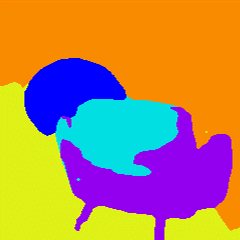}
    \includegraphics[width=1\linewidth]{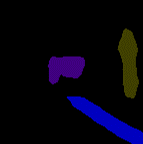}
    \includegraphics[width=1\linewidth]{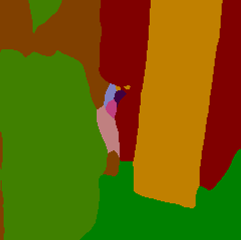}
    \caption{Base}
  \end{subfigure}
  \begin{subfigure}{0.19\linewidth}
    \includegraphics[width=1\linewidth]{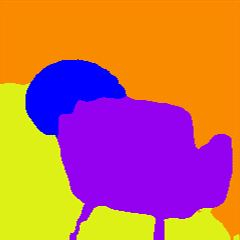}
    \includegraphics[width=1\linewidth]{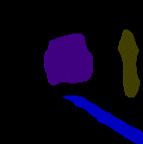}
    \includegraphics[width=1\linewidth]{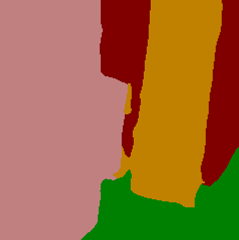}
    \caption{Ours}
  \end{subfigure}
   \caption{Visual comparisons of Base~{\small(\textit{w/o} SMMCL)}~and Ours {\small(\textit{w} SMMCL)}
   on low-light, nighttime, and normal-light scenes.  % SegNeXt-B
   }  
   \label{fig9_wmmcl}
\end{figure}

\textbf{Visual Comparisons of Base and Ours.}  
As a more intuitive validation, 
we compare Base~{\small(\textit{w/o} SMMCL)} and Ours~{\small(\textit{w}~SMMCL)} in~\cref{fig9_wmmcl}.
Thanks to our multi-modal contrastive learning approach, our model can predict different semantic classes more accurately and understand dark scenes from multi-modal images more effectively. % 4.2 4.3 4.4 

\textbf{Comparisons with Other Approaches.} 
Since our approach is the first supervised multi-modal contrastive learning approach for image segmentation, 
we comprehensively compare it with an unsupervised multi-modal approach~\cite{iros20_unirecognition} and a supervised single-modal approach~\cite{iccv2021_region}. 
% 踩supervised和multi-modal两个点
Unlike these methods which need to generate samples via augmentation or consider only single-modal region features, we leverage class labels to 
effectively align pixel embeddings across different modalities.
\Cref{table6_comparison} shows that our approach significantly outperforms them on various tasks.
This demonstrates again our effectiveness,  
and justifies the superiority of our supervised paradigm and the benefit of fully considering the cross-modal context-geometry correspondence.

\textbf{Broader Significance.} 
In our tasks, the adoption of our approach can help overcome a learning bias problem caused by ``invalid'' auxiliary modality. 
We provide additional discussions in the supplementary material.

\begin{table}[t] 
\begin{center}
\small
\begin{tabular}{|c|c|c|c|c|c|}
\hline
Method                             & S & MM & Low   & Night& Normal    \\ \hline \hline
\multicolumn{3}{|c|}{~}     & \multicolumn{3}{c|}{mIoU (\%)}                   \\ \hline 
Base                               & -                  & -                  & 66.02 & 58.44 & 54.70    \\ \hline
Base +~\cite{iros20_unirecognition}&{\small \ding{55}}  &\textbf{\checkmark} & 66.54 & 58.93 & 55.10    \\  \hline 
Base +~\cite{iccv2021_region}      &\textbf{\checkmark} &{\small \ding{55}}  & 68.02 & 59.48 & 55.46     \\ \hline
Base + {\small SMMCL}              &\textbf{\checkmark} &\textbf{\checkmark} & \textbf{68.76}& \textbf{60.00} & \textbf{55.77} \\ 
\hline
\end{tabular}
\end{center}
\vspace{-0.25cm}
\caption{
Comparisons with other contrastive learning approaches.
Base is the baseline, Base~{\small (\textit{w/o} SMMCL)}.  % \textit{i.e.,} 
S denotes supervised. MM denotes multi-modal.
}  
\vspace{-0.35cm}
\label{table6_comparison} 
\end{table}

%------------------------------------------------------------------------
%------------------------------------------------------------------------
 
\section{Conclusions} 

We tackle dark scene understanding by contrasting visible and auxiliary images with limited semantic information.
We propose a supervised multi-modal contrastive learning approach to boost the learning on the two modalities and encourage them to be semantic-discriminative in the feature space.
We demonstrate the effectiveness, generalizability, and applicability of our approach on low-light indoor scenes, nighttime outdoor scenes, normal-light scenes, and different image modalities.
We believe our work will contribute to dark scene semantic segmentation, which is a challenging but important task in life, and can inspire further progress in multi-modal scene understanding.

%------------------------------------------------------------------------
%------------------------------------------------------------------------

% tables 1-5 and figure 8 vspace, table 6 two vspace

\vspace{0.15cm}
\noindent\textbf{Acknowledgements.}
% XD was supported by the RIKEN Junior Research Associate (JRA) Program.  
% NY was was supported by JST, FOREST Grant Number JPMJFR206S, Japan. 
This work was supported by the RIKEN Junior Research Associate (JRA) Program and JST, FOREST under Grant Number JPMJFR206S.

%------------------------------------------------------------------------
%------------------------------------------------------------------------

%%%%%%%%% REFERENCES
% {\small
% \bibliographystyle{ieee_fullname}
% \bibliography{egbib}
% }

% 231118, for preprint
{\small
\bibliographystyle{ieee_fullname}

\end{document}